\documentclass[journal]{IEEEtran}

\ifCLASSINFOpdf
\else
\fi

\usepackage{graphicx}
\usepackage{caption}
\usepackage{multirow}
\usepackage{color}
\usepackage[table]{xcolor}
\usepackage{hyperref}
\begin{document}

\title{Extractive Multi Document Summarization using Dynamical Measurements of Complex Networks}

\author{Jorge Valverde Tohalino, Diego Raphael Amancio \\ Institute of Mathematics and Computer Science \\ University of S\~ao Paulo \\ S\~ao Carlos, SP, 13566-590,
Brazil \\ Email: andoni.valverde@usp.br , diego.raphael@gmail.com}

\maketitle

\begin{abstract}
Due to the large amount of textual information available on Internet, it is of paramount relevance to use techniques that find relevant and concise content. A typical task devoted to the identification of informative sentences in documents is the so called extractive document summarization task. In this paper, we use complex network concepts to devise an extractive Multi Document Summarization (MDS) method, which extracts the most central sentences from several textual sources. In the proposed model, texts are represented as networks, where nodes represent sentences and the edges are established based on the number of shared words. Differently from previous works, the identification of relevant terms is guided by the characterization of nodes via dynamical measurements of complex networks, including symmetry, accessibility and absorption time. The evaluation of the proposed system revealed that excellent results were obtained with particular dynamical measurements, including those based on the exploration of networks via random walks.
\end{abstract}


\IEEEpeerreviewmaketitle
\section{Introduction}
%


The large amount of information generated every single day has motivated the creation
of automatic methods to classify, understand and present the information in a clear and concise way.
Automatic summarization techniques are one of the many solutions to address the
problem of managing large volumes of data. Such methods aim at creating a compressed version of one
or more texts by extracting their most important content~\cite{ferreira}.

Automatic summarization techniques are divided into two groups: extractive summarization and abstractive summarization.  Extractive summaries
are produced by concatenating several sentences. Such sentences are selected
exactly as they appear in the original document. On the other hand, abstractive
summarization is a more difficult task since it includes paraphrasing sections
of the source document. In addition, abstractive methods may reuse clauses
or phrases from original documents~\cite{Nenkova:2011}.
In this paper, we focus our analysis on the extractive version of summarization techniques. Traditional techniques to select relevant sentences include the analysis of word frequency, sentence clustering and machine learning~\cite{nenkova2012survey}. Of particular interest to the aims of this paper are the methods based on complex networks.

In recent years, studies in complex networks have drawn enormous attention,
since networked models have been useful to model several real-world phenomena.
Complex networks are graphs with certain statistical and topological properties which are not common in simple and random graphs~\cite{lucas}. These properties are observed in small- world, scale-free and modular networks~\cite{watts1998},~\cite{albert2002statistical},~\cite{VIANA2013371},~\cite{PhysRevE.70.066111},~\cite{0295-5075-99-4-48002}.

Complex network concepts have proven suitable to analyze texts in several applications~\cite{0295-5075-98-1-18002,10.1371/journal.pone.0067310,1742-5468-2015-3-P03005}, including those devoted to create informative extractive summaries from one or more documents~\cite{Ribaldo2012}. Such networks can capture text structure in several ways. Nodes can represent words, sentences or paragraphs of a document and the edges between nodes are established in different ways. According to some particular measurements, nodes (e.g. sentences) receive a relevance score, which in turn is used to select as a criterion to select a subset of the best ranked sentences to compose the final extract.

In our method, with the aim of making a summary from a set of documents on the same topic (MDS), we represent nodes as sentences and network edges are established according to a similarity based on the number of shared terms between two sentences. In addition to the traditional network measurements, we used novel dynamical measurements to improve the characterization of the obtained complex networks. The summaries were produced for the CSTNews corpus~\cite{Ribaldo2012}, which comprises documents in Brazilian Portuguese. The evaluation was carried out by using the ROUGE-1 metric~\cite{Rouge}. Here we show that informative sentences can be retrieved via dynamical network measurements based on random walks, as revealed by the excellent performance
obtained mostly with measures reflecting the dynamical behavior of complex networks. The most prominent dynamical measurements were the accessibility, the absorption time and the PageRank.


This paper is organized as follows: Section \ref{section:related} contains a brief survey of works that use complex networks for extractive summarization. In Section \ref{section:metodologia}, we
detail the methodology, which includes a description of the proposed network model and the networks measurements used to select the best ranked sentences to compose the summary. In Section \ref{section:results}, the results are presented and discussed. Finally, the conclusions and prospects for future work are shown in Section \ref{section:conclusions}.

\section{Related Work}\label{section:related}
Several works addressing extractive summarization based on graphs
and complex networks measurements have been proposed. In the work of Antiqueira et al.~\cite{lucas}, nodes represent sentences and an edge connect two sentences if they share significant lemmatized nouns. Then, in order to give a numerical
value to each node, some complex network measurements are applied. The best
ranked nodes (sentences) are selected to compose the final extract. Antiqueira et al. \cite{lucas} also implemented a summarizer based on a voting system, which combines the results of summaries generated by different measurements. The TeM\'ario corpus was used~\cite{lucas} to evaluate the results. Some of the proposed systems achieved similar results compared to the top summarizers for Brazilian Portuguese.

Leite and Rino \cite{Leite} explored multiple features using machine learning.
The authors took into account $11$ SuPor-v2 \cite{LeiteR06} features, which is a
supervised summarizer for Brazilian Portuguese, and 26 features based on
complex network measurements. In order to compose the extract using a machine learning perspective, each sentence was classified as present or not present in summary. Leite and Rino \cite{Leite} also used TeM\'ario corpus to evaluate the results, obtaining excellent results.

Ribaldo et al.~\cite{Ribaldo2012} addressed the Multi Document Summarization
(MDS) task for texts in Brazilian Portuguese. All sentences from the corpus were
identified and modeled as a single network. The pre-processed sentences
were represented as nodes, which were linked by similarity measurements.
To select the best ranked sentences, the authors used
the degree, clustering coefficient and the shortest paths measurements. To create a summary devoid of redundancy, the authors proposed a method to remove
sentences with same content. The evaluation of results was performed on
the CSTNews Brazilian corpus. The reported results showed that their method for Portuguese MDS yielded very good results, which were close to the best
system available for the language.

In the work of Amancio et al.~\cite{diego}, networks are created as follows:
each lemmatized word is represented as a single node and edges are obtained by
joining nodes whose corresponding words are immediately adjacent in the text. Each edge
weight is determined by the number of repeated associations between two words.
After building a network, several measurements (strength, shortest paths, betweenness,
vulnerability and diversity) are computed at the word level. For sentence selection, each sentence receives a weight based on the average weight of its content words. Finally, the
$n$ best weighted sentences are included in the final summary. The authors found that diversity-based measurements outperformed the best system proposed in~\cite{lucas}.

In the work of Salton et al.~\cite{Salton}, text paragraphs are represented as
nodes and edges are established between two nodes according to a similarity measure
based on the number of shared words. Routing algorithms, such as bushy
and depth first paths were used to select the most
important paragraphs. The algorithms were evaluated using a corpus of 50
documents in English. The best algorithm selected 45.6\% of paragraphs chosen by
human summarizers.

Mihalcea~\cite{Mihalcea:2005} defined a network of sentences which are connected
according to the number of terms they share. To select
the most informative sentences, Mihalcea used recommendation algorithms for Web
Pages, including both Google's Page Rank \cite{Pageetal98} and HITS
\cite{Kleinberg:1999}. Three network types were considered: undirected,
forward (edges reflecting the reading flow of text) and backward (edges going from the current
to the previous word in the text). The evaluation was performed on the English
corpus DUC'2002 \cite{Over2002} and the Portuguese corpus TeM\'ario
\cite{lucas}. The results of the
HITS algorithms were superior to the best DUC'2002 system when both forward and
backward networks were used. For the Portuguese scenario, the Backward network evaluated
by the Page Rank algorithm provided the best performance.


\section{Methodology}\label{section:metodologia}

In the current paper, we propose a method based on complex network measurements for
Portuguese Multi Document Summarization (MDS). We make an extension of Antiqueira et
al. \cite{lucas} and Ribaldo et al. \cite{Ribaldo2012} works by using new dynamical network
measurements to characterize complex networks. Each extracted sentence from
documents is represented by a node and the edges are created if two sentences are
semantically similar. This proposal is divided into five stages: Document pre-processing, sentence vectorization, network creation, application of network measurements and summarization (i.e., sentence selection).

\subsection{Document pre-processing}

In order to model sentences as network nodes, a set of changes must be
applied to the original texts. Such changes include the elimination of unnecessary
words and the transformation of words into their canonical form. This stage includes:
\begin{itemize}

  \item Text segmentation: this stage divides texts into sentences. We consider as a
  sentence any text segment separated by a period, exclamation or question
  mark. We used the Python NLTK library~\cite{nltk2006} for the text
  segmentation.

  \item Elimination of stopwords and punctuation marks: For the elimination of
  unnecessary words, we used a list of stopwords for Portuguese.

  \item Morphosyntactic labeling: Part Of Speech Tagging is important for word
  lemmatization and for the identification of all nouns composing a
  sentence. In this phase, we used the MXPost Tagger~\cite{tagger} for
  Portuguese.

  \item Lemmatization: In this phase, we obtained the canonical form of each word with the aim of processing in the same canonical form different variations of a word.

\end{itemize}
Table \ref{tab:procesamiento} shows an example of the document pre-processing
stage.

\begin{table*}[ht]
\centering
\caption{Example of the pre-processing stage applied to a piece of text
extracted from Wikipedia. The first column shows the original text divided into six
sentences. In the second column, we show the pre-processed text and shared nouns
between sentences are highlighted.}
\label{tab:procesamiento}
\begin{tabular}{ll}
\hline
\multicolumn{1}{c}{\textbf{Original text divided into sentences}} &
\multicolumn{1}{c}{\textbf{Pre-processed text}}
\\
\hline 1. Brazil is the largest country in South America &
\textbf{brazil} be large \textbf{country} \textbf{south} \textbf{america}
\\

2. It is the world's fifth-largest country by both area and population &
be \textbf{world} five large \textbf{country} area population \\

3. It is the largest country to have Portuguese as an official language and the
only one in America & be large \textbf{country} have portuguese
official language \textbf{america} \\

4. Bounded by the Atlantic Ocean on the east, Brazil has a coastline of 7,491
kilometers & bound atlantic ocean east \textbf{brazil} have coastline
kilometer \\

5. It borders all other South American countries except Ecuador and Chile &
border \textbf{south} \textbf{america} \textbf{country} ecuador chile \\

6. Brazil's economy is the world's ninth-largest by nominal GDP of 2015 &
\textbf{brazil} economy be \textbf{world} nine large nominal gdp \\
\hline
\end{tabular}
\end{table*}

\subsection{Sentence vectorization}
We used the Tf-Idf weighting for vector representation of sentences since this
metric was employed with satisfactory results for many NLP tasks
\cite{robertson2004}. To get the vector of a sentence, we calculate the Tf-Idf
value of each of its corresponding words, where Tf is the term frequency and Idf is the inverse document frequency.


\subsection{Network creation}

\begin{figure}[ht]
 \centering
 \includegraphics[scale=0.26]{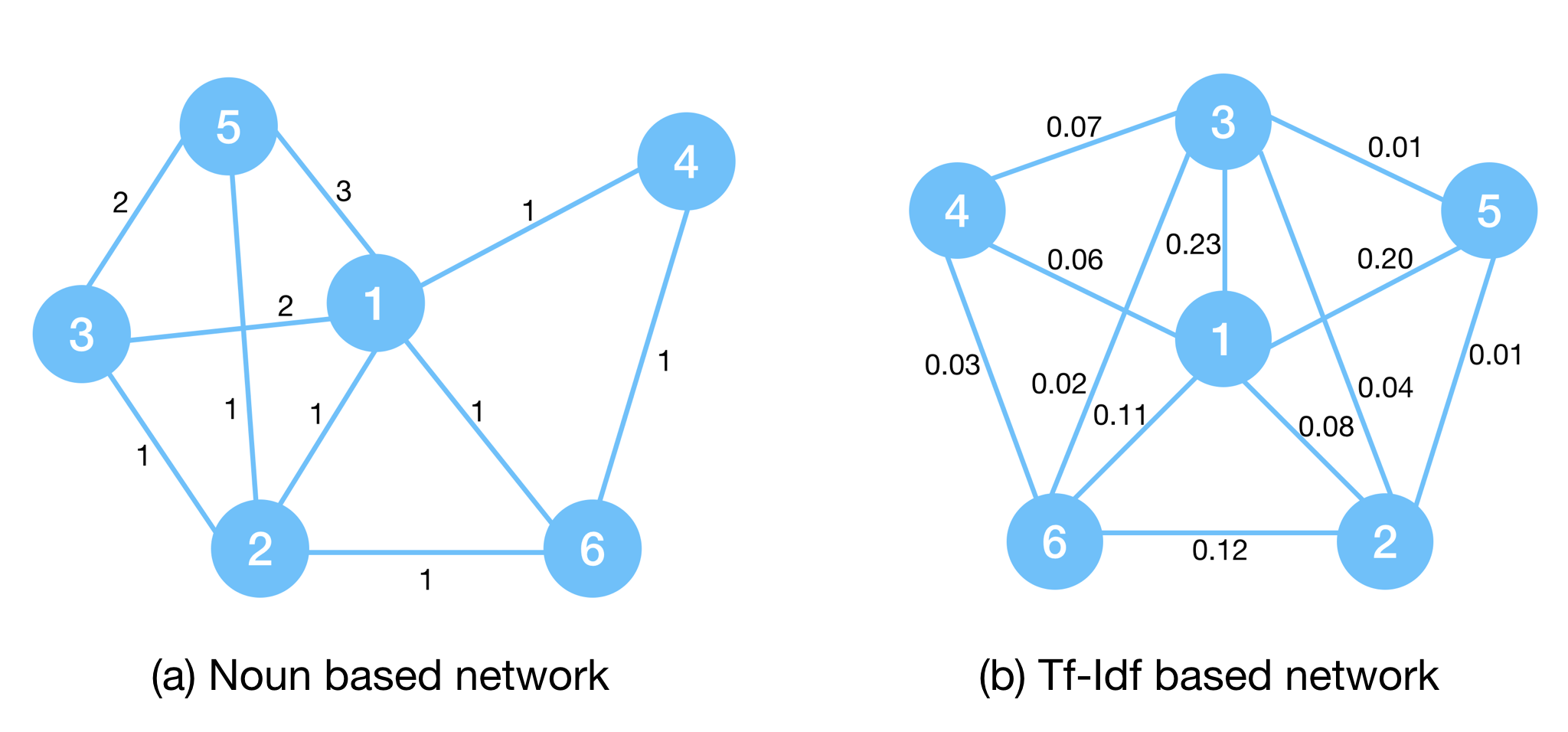}
 \caption{Network models proposed in this work.}
 \label{fig:networks}
\end{figure}

This stage creates two network models for document
representation. The first network, hereafter referred to as Noun based network, follows the
Antiqueira's work~\cite{lucas}. The second variation of network, hereafter referred to as Tf-Idf based network, is based on Ribaldo's work~\cite{Ribaldo2012}. The particularities of these models are:

\begin{itemize}

  \item Noun based network: In this model, each node represents a sentence comprising  lemmatized nouns. There is an edge between two sentences when there is at least one noun in common between such sentences. The number of word repetitions between both sentences indicates the edge weight linking the sentences.

  \item Tf-Idf based network: To create this network, we first need to
  determine the Tf-Idf vector representation of each document sentences. Then,
  each node network is represented by a sentence and the edge between two
  sentences is based on the similarity between the Tf-Idf vectors of both
  sentences. The similarity is computed as the cosine similarity obtained
  from the Tf-Idf vectors.
\end{itemize}
Figure \ref{fig:networks} shows an example of the two
network models proposed in this work, which were generated from the example in
Table \ref{tab:procesamiento}.

\subsection{Application of network measurements}
In this stage, we use a set of network measurements with the aim of giving a
value of importance (weight) to each node. This weight allows us to rank the
nodes, so that the the best ranked sentences (nodes) compose the final
summary. {Every network measurement is used in an individual way, therefore, there is one summary for each measurement.} In addition to the traditional network measurements (degree, strength, shortest paths, clustering coefficient, betweenness, and page rank), we
used additional measurements to take into account not only
the topological structure of the networks, but also their dynamical behavior.
This can be achieved by considering dynamical processes occurring on the top
of the networks. For simplicity sake, we considered variation of random walks
to study the dynamical behavior of the networks. Such a dynamics gives rise to
a series of measurements, including concentric metrics,
accessibility, symmetry and absorption time. The metrics employed in this work
are detailed below.


\begin{itemize}
  \item Degree: The degree of a vertex $i$ is the number of edges
  connected to that vertex. 
  \item Strength: For weighted networks, the strength of a
  vertex $i$ is the sum of the weights of its corresponding edges. 
  \item Clustering coefficient: It is a measure to characterize the
  presence of loops of order three in a network. It measures the probability that two neighbors of a node are connected.
  \item Shortest paths: A shortest path between two vertices $i$ and
 $j$ is one of the paths  ($d_{ij}$) that connects these vertices with a minimum
 length. The length of a path is determined by the number of edges connecting
 the vertices $i$ and $j$. The similarity between distances is converted to distances using two distinct rules:
 \begin{itemize}
 \item $w^{(1)}_{ij}=0$ if $w_{ij}=0$ or $w^{(1)}_{ij}=w_{max}-w_{ij}+1$ if $w_{ij}>0$; and
 \item $w^{(2)}_{ij}$ = 0 if $w_{ij}=0$ or $w^{(2)}_{ij}=1/w_{ij}$  if $w_{ij}>0$
 \end{itemize}
where $w_{ij}$ is an element of the weighted matrix $W$ representing
the edges weight (i.e. the original similarity indexes) and $w^{(1)}$ and $w^{(2)}$ are the obtained distances.

  \item Betweenness: This measure is normally calculated as the
  fraction of shortest paths between two nodes passing through some node of
  interest. 
  \item Page Rank: In  this measurement, a node $i$ is relevant if it is connected to other relevant nodes.


  \item Concentric measurements: this type of measurement represents a set of eight indexes that are able to extract valuable topological information along hierarchical levels of networks \cite{concentrics}.  A hierarchical level allows a natural and powerful extension of basic measurements. The basic definition of any concentric measure relies on the identification of  the  ring  $R_d(i)$, i.e. the set of nodes which are $d$ hops away from $i$. The following are some of the metrics that were proposed in the work of Costa and Silva \cite{concentrics}:
  \begin{enumerate}
    \item  Concentric number of nodes: Number of nodes belonging to the
  ring  $R_d(i)$.
    \item Concentric number of edges: Number of edges connecting nodes inside
    the ring $R_d(i)$.
    \item Concentric node degree: Number of edges extending from the ring
  $R_d(i)$ to $R_{d+1}(i)$.
    \item Concentric clustering coefficient: Division of the number of
  existing edges in the ring $R_d(i)$ by the total number of possible edges in
  this ring.
    \item Convergence ratio: Ratio between the concentric node degree
  and the number of nodes at the next concentric ring.
    \item Intra-ring node degree: It is the average degree of the nodes
  at the ring $R_d(i)$ considering only the edges located in the ring $R_d(i)$.
    \item Inter-ring node degree: It is the ratio between the node
  degree and the number of nodes in the same ring.
    \item Concentric common degree: The average degree considering all
    the connections of nodes at a specific ring.
  \end{enumerate}
  \item Accessibility: The accessibility quantifies the number of nodes
  actually accessible from an initial node \cite{accesibilidad}. To calculate
  this measure, consider that $P^{(h)}(i,j)$ represents the probability of reaching
  a vertex $j$ from $i$ through a self-avoiding random walk of length $h$
  \cite{accesibilidad}. This measure considers the paths from the vertex $i$ to
  each of the vertices located in the concentric ring of distance $h$, and it is
  calculated as
  \begin{equation}
  k_{i}(h) = \exp \Bigg(-\sum_{j} P^{(h)}(i,j) \log P^{(h)}(i,j)\Bigg)
  \end{equation}
  \item Generalized accessibility: because  the accessibility measurement depends
  on the parameter $h$, a new version of accessibility can be considered without such a   parameter. The generalized accessibility is based on a Matrix Exponential Operation.  This operation allows the calculation of transition probabilities considering walks of \emph{all lengths} between any pair of vertices. This measurement has been employed with sucess in other text classification tasks~\cite{simetria}.

\item Symmetry: The network symmetry is a normalized version of
  accessibility, where the number of accessible nodes is used as normalization
  factor \cite{simetria}. To calculate this measure, concentric random walks
  are used as a way to avoid transitions to previous concentric levels.
  Therefore, changes must be made in the network so that the transitions do not
  use edges within a same concentric level. These changes originate two types of
  symmetry: backbone symmetry and merged symmetry. {In the backbone symmetry, the edges that connect nodes belonging to the same concentric level are disregarded. In the merged symmetry, these edges have cost 0 and the nodes connected by them are collapsed.}
  The symmetry is calculated as:
  \begin{equation}
  \label{eq:sym}
  S_i^{(h)} = \frac{\exp(-\sum P^{(h)}(i,j) \log P^{(h)}(i,j))}{| \xi_i^{(h)}|},
  \end{equation}
  {where $P^{(h)}(i,j)$ is the probability of reaching a node $j$ from node $i$ through a self-avoiding random walk of length $h$}, and $\xi_i^{(h)}$ is the set of accessible nodes that are at a distance $h$
  from the node $i$. The objective of using this measurement is to determine
  if nodes with a higher degree or lower degree of symmetry are good indicators
  of sentence importance. We tested this metric by selecting nodes with greater
  and lower symmetry.

  \item Absorption Time: This metric is defined as the time it takes for a particle
  in an internal node to reach an output node through a random walk. The
  absorption time $\tau$ quantifies how fast a randomly-walking particle is
  absorbed by the output vertices, assuming that the particle starts the random
  walk at the input node~\cite{absortion}. According to this measurement,
  sentences with lower absorption time are probably the appropriate sentences
  to form part of the summary.

\end{itemize}
Table \ref{tab:medidas} summarizes the adopted network measurements for this work
and how they are going to be used for summarization purposes.
\begin{table}[ht]
\scriptsize
\centering
\caption{Adopted network measurements for MDS. The weighted version of the networks was considered only with the traditional measurements.}
\label{tab:medidas}
\begin{tabular}{|c|c|c|c|}
\hline
\multirow{2}{*}{\textbf{Selection}}                                              & \multirow{2}{*}{\textbf{Measurement}} & \multirow{2}{*}{\textbf{Abbr.}} & \multirow{2}{*}{\textit{\textbf{\begin{tabular}[c]{@{}c@{}}Hier. level\\ h=2/h=3\end{tabular}}}} \\ & & &                                                                                            \\ \hline
\multirow{9}{*}{\begin{tabular}[c]{@{}c@{}}Highest\\ values\\ \end{tabular}}
& Degree & Dg &                                                                                     \\ 
                                                                                 & Strength & Stg &                                                                                                  \\ 

& Betweenness & Btw/Btw\_w &                                                                                                   \\ 
                                                                                 & Page Rank & PR/PR\_w &                                                                                                  \\ 
                                                                                 & Clustering Coefficient & CC/CC\_w&                                                                                                  \\ 
                                                                                 & Concentric & Conc\_\{1,...,8\}& x                                                                                                \\ 
                                                                                 & Symmetry & HSymBb/HSymMg& x                                                                                                \\ 
                                                                                 & Accessibility & Access& x                                                                                                \\ 
                                                                                 & Generalized Accessibility & GAccess&                                                                                                  \\ \hline
\multirow{3}{*}{\begin{tabular}[c]{@{}c@{}}Lowest\\ Weighted\\ Nodes\end{tabular}}
& Shortest Paths & SP/SP\_w1/SP\_w2 &                                                                                                  \\ 
                                                                                 & Symmetry & LSymBb/LSymMg& x                                                                                                \\ 
                                                                                 & Absorption Time & AbsT&                                                                                                  \\ \hline
\end{tabular}
\end{table}

\subsection{Summarization}
In this stage, the best ranked sentences are selected to compose the
summary. In the first place, generated summaries must respect a established size. This
size is adapted according to the size of references summaries. Generally summaries have
a compression rate of 70\% of the original text \cite{Ribaldo2012}.

Also, for MDS, it is important to avoid redundancy in the selected
sentences. Redundancy could occur when identical or similar sentences are
represented in the graph as different nodes and it is frequently indicated by
links with very high degree of similarity \cite{Ribaldo2012}. In this paper, we
use the anti-redundancy method proposed by Ribaldo et al. \cite{Ribaldo2012}. In
this work, it is set a redundancy limit that a new selected sentence may have
in relation to any of the previously selected sentences. If this limit is
reached, this new sentence is considered redundant and it is ignored, and the
summarization process goes to the next candidate sentence; otherwise, the
sentence is included in the summary. Ribaldo et al. \cite{Ribaldo2012} defined
this limit as the sum of the highest and the lowest cosine similarity between
all sentences of the original texts.

\section{Results}\label{section:results}
In this section, we show the achieved results from the evaluation of our systems
for Portuguese Multi Document Summarization (MDS). We used the CSTNews corpus
\cite{Ribaldo2012}, which is a set of documents that were extracted from on-line
Brazilian news agencies. This corpus contains 140 news items, which are divided
into 50 clusters. Each cluster contains 2 or 3 documents on the same
topic. Our systems were evaluated by using the ROUGE-1 metric \cite{Rouge},
which compares the generated summaries and the human-generated summaries from
CSTNews. This metric was used because, it has been shown that there is a strong correlation between ROUGE indexes and manual (human) judgement.
For comparison purposes, Table \ref{tab:others} shows the results from other
works that achieved the best results for MDS: GistSumm \cite{gist}, which was
the first MDS system produced for Portuguese; CSTSumm \cite{cst2}, which follows a
CST-based method (cross-document Structure Theory); MEAD \cite{mead}, that is
based on centroids, sentence position and lexical features extracted from the
sentences; and BushyPath and Depth-first Path systems~\cite{Ribaldo2012}, which
adapt the Relationship Map approach for MDS.
\begin{table}[h]
\centering
\caption{List of works for Portuguese MDS with the respective average ROUGE-1 Recall (RG1) scores.}
\label{tab:others}
\begin{tabular}{cccc}
\hline
\textbf{Systems} & \textbf{RG1} & \textbf{Systems} & \textbf{RG1}
\\
\hline GistSumm & 0.6643  & MEAD & 0.4602  \\
BushyPath & 0.5397 & Top Baseline & 0.5349  \\
Depth-first Path & 0.5340 & Random Baseline  & 0.4629  \\
CSTSumm & 0.5065 &  &  \\  \hline
\end{tabular}
\end{table}
In this paper, in order to compare our systems with other works shown in Table \ref{tab:others} and two baselines, we show in Table \ref{tab:results} the average ROUGE-1 Recall scores obtained from the proposed systems. With the aim of generating baseline summaries, the first baseline, called Top Baseline, selects the first $n$ sentences of the source document, while the Random Baseline randomly selects sentences from the source document~\cite{lucas}.
\begin{table}[ht]
\centering
\caption{{RG1-R} results for Portuguese MDS. In in the first two columns is shown
the performance of our systems without using the anti-redundancy detection
method (ARD). In last columns is shown the results by using the anti-redundancy
detection method (ARD). Results in blue represent the 5 best systems,
while those in orange represent the 5 worst systems.}
\label{tab:results}
\begin{tabular}{c|c|c|c|c|}
\cline{2-5}
\multicolumn{1}{l|}{} & \multicolumn{2}{c|}{\textbf{MDS}} &
\multicolumn{2}{c|}{\textbf{MDS + ARD}} \\ \hline
\multicolumn{1}{|c|}{\textbf{Measures}} & \multicolumn{1}{l|}{\textbf{Noun}} &
   \multicolumn{1}{l|}{\textbf{Tf-Idf}}
&  \multicolumn{1}{l|}{\textbf{Noun}}
&   \multicolumn{1}{l|}{\textbf{Tf-Idf}}
 \\ \hline

\multicolumn{1}{|c|}{Dg} & \cellcolor{cyan!25}0.5469 & \cellcolor{cyan!25}0.5482& 0.5400 & \cellcolor{cyan!25}0.5528\\
\multicolumn{1}{|c|}{Stg} &\cellcolor{cyan!25}0.5453 & 0.5390 &\cellcolor{cyan!25} 0.5433&\cellcolor{cyan!25}0.5552 \\
\multicolumn{1}{|c|}{SP} & 0.5441 & 0.5438 & 0.5432 & 0.5509  \\
\multicolumn{1}{|c|}{SP-w1} & 0.5346 & \cellcolor{cyan!25}0.5478 &\cellcolor{cyan!25} 0.5454&\cellcolor{cyan!25}0.5636 \\
\multicolumn{1}{|c|}{SP-w2} & 0.5417 & 0.5314 & \cellcolor{cyan!25} 0.5545 &0.5515 \\
\multicolumn{1}{|c|}{Btw} & 0.5298 & 0.5404 & 0.5341 & 0.5452\\
\multicolumn{1}{|c|}{Btw-w} & 0.4763 & 0.4745 & 0.4901 & 0.4790 \\
\multicolumn{1}{|c|}{PR} & \cellcolor{cyan!25}0.5501 & 0.5367 & 0.5426 &0.5435\\
\multicolumn{1}{|c|}{PR-w} & \cellcolor{cyan!25}0.5458 &\cellcolor{cyan!25}0.5460 & \cellcolor{cyan!25} 0.5471&\cellcolor{cyan!25}0.5605\\
\multicolumn{1}{|c|}{CC} & 0.4151 & 0.4266 & 0.4270 & 0.4424\\
\multicolumn{1}{|c|}{CC-w} & 0.4180 & 0.4326 & 0.4337 & 0.4532\\
\multicolumn{1}{|c|}{Conc-1(h=2,3)} & \cellcolor{orange!25}0.3999 &\cellcolor{orange!25}0.3957 & \cellcolor{orange!25} 0.4171&\cellcolor{orange!25}0.4083 \\
\multicolumn{1}{|c|}{Conc-2(h=2,3)} & \cellcolor{orange!25}0.3943 &\cellcolor{orange!25}0.3895 & \cellcolor{orange!25} 0.4157&\cellcolor{orange!25}0.4057\\
\multicolumn{1}{|c|}{Conc-3(h=2,3)} & 0.4035 & \cellcolor{orange!25}0.4095 &0.4246 & \cellcolor{orange!25}0.4187\\
\multicolumn{1}{|c|}{Conc-4(h=2,3)} & \cellcolor{orange!25}0.3919 &\cellcolor{orange!25}0.3858 & \cellcolor{orange!25} 0.4115&\cellcolor{orange!25}0.4068 \\
\multicolumn{1}{|c|}{Conc-5(h=2,3)} & 0.4204 & 0.4214 & 0.4376 & 0.4324 \\
\multicolumn{1}{|c|}{Conc-6(h=2,3)} & 0.4077 & 0.4259 & 0.4235 & 0.4393  \\
\multicolumn{1}{|c|}{Conc-7(h=2,3)} &\cellcolor{orange!25}0.3989 &\cellcolor{orange!25}0.3730 & \cellcolor{orange!25} 0.4116&\cellcolor{orange!25}0.3934  \\
\multicolumn{1}{|c|}{Conc-8(h=2,3)} &0.4179 & 0.4276 & 0.4283 & 0.4432 \\
\multicolumn{1}{|c|}{Access (h=2)} &0.4925 & 0.5093 & 0.5032 & 0.5102 \\
\multicolumn{1}{|c|}{Access (h=3)} & 0.4484 & 0.4302 & 0.4540 & 0.4369  \\
\multicolumn{1}{|c|}{GAccess} & \cellcolor{cyan!25}0.5489 &\cellcolor{cyan!25}0.5478 & 0.5395 & 0.5494   \\
\multicolumn{1}{|c|}{HSymBb (h=2)} & 0.4183 & 0.4202 & 0.4242 & 0.4228  \\
\multicolumn{1}{|c|}{HSymBb (h=3)} & \cellcolor{orange!25}0.4010 & 0.4307 &\cellcolor{orange!25} 0.4200 & 0.4438\\
\multicolumn{1}{|c|}{HSymMg (h=2)} & 0.4745 & 0.4856 & 0.4829 & 0.4906  \\
\multicolumn{1}{|c|}{HSymMg (h=3)} & 0.4525 & 0.4621 & 0.4591 & 0.4744  \\
\multicolumn{1}{|c|}{LSymBb (h=2)} & 0.5207 &0.5302& 0.5288 & 0.5461 \\
\multicolumn{1}{|c|}{LSymBb (h=3)} & 0.4829 & 0.4716 & 0.4918 & 0.4732  \\
\multicolumn{1}{|c|}{LSymMg (h=2)} & 0.4576 & 0.4731 & 0.4712 & 0.4763  \\
\multicolumn{1}{|c|}{LSymMg (h=3)} & 0.4780 & 0.4664 & 0.4896 & 0.4725   \\
\multicolumn{1}{|c|}{AbsT} &0.5435 & \cellcolor{cyan!25}0.5449 &\cellcolor{cyan!25} 0.5441&\cellcolor{cyan!25}0.5534\\ \hline
\end{tabular}
\end{table}

In this work, two experiments were carried out. In the first approach, we make a simple selection of best ranked sentences without using the anti-redundancy detection method (ARD). In the second approach, the ARD method is used.
The results in Table \ref{tab:results} show that applying anti-redundancy detection (ARD) methods does not have a big impact on the summary quality. We can see that ARD methods had a slightly better performance than the simple sentence selection method. In some cases, the results obtained without the ARD method outperformed the ones obtained with such a filtering of sentences. We could conclude there is not great relevance in applying the adopted ARD methods for the CSTNews corpus. It remains, therefore, to be probed in future works the efficiency of other methods for elimination of redundant sentences. 

The proposed methods achieved a good performance since they outperformed the majority of the results from other works for MDS. We evaluated the Noun and Tf-Idf based networks. Both networks displayed a similar performance. According to Table \ref{tab:results}, traditional network measurements like degree, shortest paths, page rank, betweenness, and some of their weighted versions yielded the best scores.  The measurements based on the dynamical behavior of the networks, such as absorption time and generalized accessibility measurements also displayed an excellent performance. The backbone symmetry measurement ($h=2)$ achieved a good performance when the least symmetric nodes were taken into account; in other cases, however, symmetry measurements yielded very low ROUGE scores. The accessibility measurement was outperformed by the top baseline score, when it was evaluated at the second hierarchical level. Such a performance decreased when further hierarchical levels were taken into account. Finally, the systems based on concentric and clustering measurements yielded the lowest results. 


\section{Final remarks}\label{section:conclusions}
In this paper, we probed the efficiency of several complex networks measurements for the multi document extractive summarization task.
We used novel dynamical complex networks metrics, such as absorption time and generalized accessibility, which achieved excellent scores. Our results suggest that such measurements could be used to improve the characterization of networks for the summarization task, as they complement the traditional analysis using a dynamical point of view.
Because all these measurements are based on a random walk with distinct preferential strategies, we believe that such a walk should be further explored in further works.
In order to improve the summary quality, it would be important to use more sophisticated methods to represent documents as networks. These methods include, for example, word embeddings~\cite{embeddings} to get a better representation of texts. Also, in another approach, the document set could be represented as a multilayer network~\cite{boccaletti2014structure}, where each network layer corresponds to a different document of the group of documents. It would also be important to develop an approach that combines both traditional document summarization techniques and complex network concepts. For example, methods based on machine learning could be combined  with traditional features like sentence length, proper nouns or sentence location.




\section*{Acknowledgment}

The authors acknowledge financial support from CNPq, CAPES, and S\^ao Paulo Research Foundation (FAPESP grant no. 16/19069-9).

\ifCLASSOPTIONcaptionsoff
  \newpage
\fi

\bibliographystyle{IEEEtran}
\bibliography{references}


\end{document}